\documentclass[10pt,twocolumn,letterpaper]{article}

\usepackage[pagenumbers]{cvpr}      %

\patchcmd{\abstract}{\vspace*{12pt}}{\vspace*{7pt}\parskip=0pt}{}{\GenericWarning{}{preamble.tex: abstract spacing patch failed}}

\usepackage{xspace}
\usepackage{placeins}
\newcommand{\ours}{OTC\xspace}

\usepackage{colortbl}
\definecolor{tableHeader}{HTML}{E6EDF5}
\definecolor{tableGroup}{HTML}{F1F3F6}
\definecolor{tableOurs}{HTML}{FFF0E5}
\definecolor{tableRule}{HTML}{64748B}
\newcommand{\paperTableStyle}{%
  \arrayrulecolor{tableRule}%
  \setlength{\aboverulesep}{0pt}%
  \setlength{\belowrulesep}{0pt}%
  \setlength{\heavyrulewidth}{0.7pt}%
  \setlength{\lightrulewidth}{0.35pt}%
  \setlength{\extrarowheight}{0.8pt}%
}

\definecolor{cvprblue}{rgb}{0.21,0.49,0.74}
\usepackage[pagebackref,breaklinks,colorlinks,allcolors=cvprblue]{hyperref}

\def\confName{CVPR}
\def\confYear{2027}

\title{Count Once, Carry Through: A Training-Free Framework for Numerically Aligned Text-to-Video Generation}

\author{HaoYuan Ma$^{1,2,\dagger}$ \quad Haoxiang Zhao$^{3}$ \quad Tianqin Zhu$^{1}$ \quad Pinyue Hao$^{4}$\\
Li Zhang$^{2,*}$ \quad Jie Gao$^{2}$ \quad Zhiyuan Ma$^{1,*}$\\[3pt]
{\normalsize $^{1}$Huazhong University of Science and Technology \quad $^{2}$Soochow University}\\
{\normalsize $^{3}$University of Science and Technology of China \quad $^{4}$Dalian University of Technology}\\
{\small $^{*}$Corresponding authors}\\
{\small $^{\dagger}$\,Work done during HaoYuan Ma's internship at Huazhong University of Science and Technology.}}

\hypersetup{pdfauthor={HaoYuan Ma, Haoxiang Zhao, Tianqin Zhu, Pinyue Hao, Li Zhang, Jie Gao, Zhiyuan Ma},pdftitle={Count Once, Carry Through: A Training-Free Framework for Numerically Aligned Text-to-Video Generation}}
\begin{document}

\twocolumn[{%
\renewcommand\twocolumn[1][]{#1}%
\maketitle
\begingroup
    \centering
    \captionsetup{type=figure,skip=4pt}
    \includegraphics[width=0.92\textwidth,trim=28bp 70bp 14bp 14bp,clip]{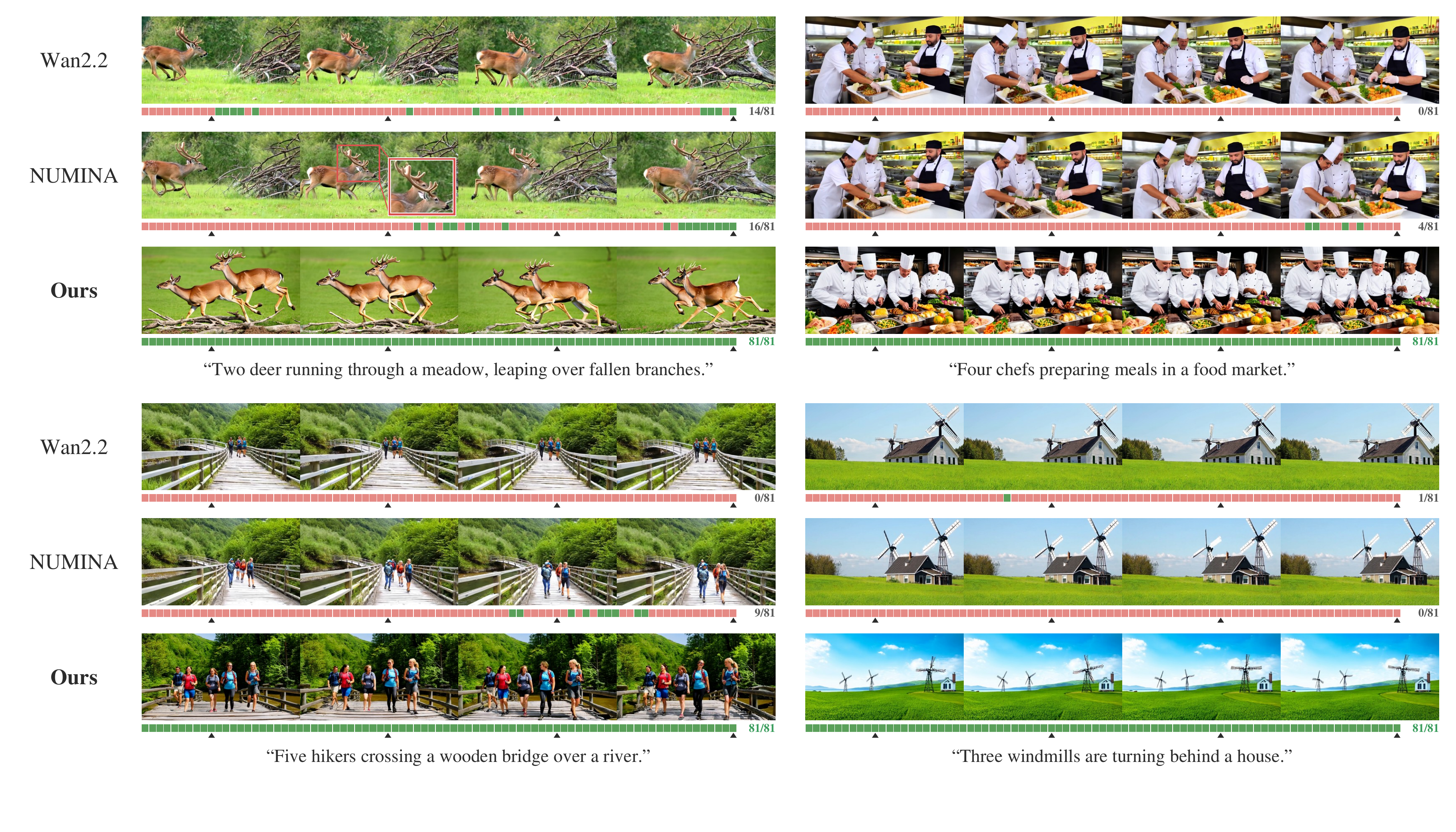}\par
    \captionof{figure}{\textbf{Qualitative comparison on CountBench with Wan2.2-T2V-5B.} The selected examples illustrate incorrect numbers of objects and temporal count drift in the base model and NUMINA~\cite{sun2026numina}. Our training-free framework, \ours, achieves the target count in all 81 frames of each example shown. The frame-wise strips cover the complete 81-frame clips.}
    \label{fig:teaser}
\endgroup
}]

\begin{abstract}
Text-to-video diffusion models often struggle to generate the specified number of objects and maintain that count throughout a video. A key challenge is that count estimates are less reliable early in denoising, whereas the global layout is already difficult to revise in later steps. We introduce \ours (Observe, Think, and Carry), a training-free framework that separates numerically aligned anchor generation from temporal propagation. Observe identifies instances in a draft anchor latent by associating self-attention clusters with object categories through cross-attention. Think performs localized edits and verifies their outcomes by recounting. Carry uses the resulting numerical anchor latent as a spatial reference during video generation. Carry Guidance (Carry-CFG) combines text guidance with the contrast between predictions conditioned on the clean anchor and its re-noised counterpart, allocating stronger anchor guidance to instance regions. Experiments on CountBench across four video diffusion backbones show average gains of 9.0 and 6.6 percentage points in counting accuracy and temporal count consistency, respectively, over the corresponding base configurations. Under the same evaluation settings, \ours also outperforms NUMINA.
\href{https://jason-mar1.github.io/OTC/}{Project demo}
\end{abstract}
    
\section{Introduction}
\label{sec:intro}

Text-to-video (T2V) diffusion models are increasingly used not only for content creation but also as world simulators and data engines~\cite{wan2025wan, kong2024hunyuanvideo, Veo3}. In many real-world applications, the number of objects is a hard constraint that must hold throughout the video. When video models simulate driving scenes or embodied tasks~\cite{GAIA1, Vista, UniPi}, the number of pedestrians, vehicles, or manipulable objects defines the scenario itself. If an object disappears unexpectedly, the simulated scenario changes, potentially misleading downstream decision-making. When generated videos are used to train detection, tracking, and counting models~\cite{TrackDiffusion}, unintended changes in object presence or number can make the videos inconsistent with their annotations. Accurate object counts are therefore essential for reliable scene simulation and synthetic training data.
However, current T2V models struggle to align the number of generated objects with the target count specified in the prompt. Figure~\ref{fig:teaser} illustrates these failures in videos produced by the base Wan2.2 model and its NUMINA-guided counterpart~\cite{sun2026numina}. These examples highlight two requirements for numerically aligned video generation: establishing the target count and preserving it throughout the video.

\begin{figure}[!t]
  \centering
  \includegraphics[width=0.95\columnwidth]{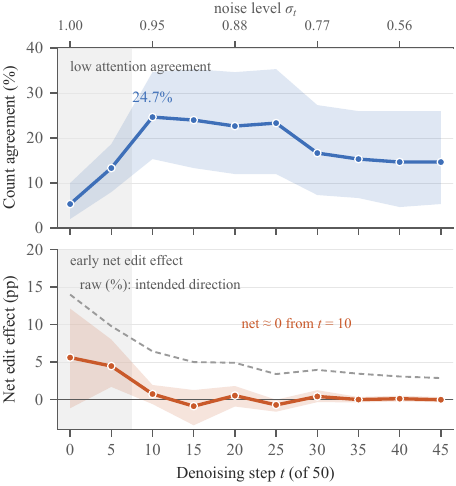}
  \captionsetup{skip=4pt}
  \caption{\textbf{Timing mismatch between count readout and attention edits.} Wan2.1-T2V-1.3B, 30 single-noun CountBench prompts. Top: attention-count agreement with final detector counts at five sampled frames. Bottom: intended minus opposite one-instance count-change rates over frames 10--81. Dashed gray shows the intended-direction rate. Both metrics are averaged within and then across prompts. Bands indicate 95\% bootstrap confidence intervals over prompts.}
  \label{fig:dilemma}
\end{figure}

Why is the number of objects difficult to control in video generation? Numerical control requires identifying instances while the layout remains responsive to editing. Early in denoising, the layout is easier to change, but individual objects remain difficult to distinguish. Toward the end of the trajectory, objects become clearer, yet their number and arrangement are harder to revise~\cite{wang2026demystifying,ATHENA,gu2026mitigating}. In our diagnostic study (\cref{fig:dilemma}), estimated net editing effects are largest at $t\le5$, whereas attention-based count agreement peaks at $t=10$, when the net effect is already close to zero. This suggests a temporal mismatch between reading the eventual count from attention and changing it through the tested edits. NUMINA~\cite{sun2026numina} builds latent layouts from selected self-attention and noun-specific cross-attention heads, then refines them with spatial and temporal costs before guiding regeneration through cross-attention modulation. In our experiments, its temporal count consistency gains vary across backbones (\cref{tab:main_results,tab:h3_results}).

To alleviate this dilemma, we separate numerically aligned anchor generation from temporal count control. Completing a single-frame latent allows its instances to be counted, edited, and verified before the remaining video is generated. Drawing on advances in object insertion, removal, and inpainting~\cite{PromptToPrompt,brooks2023instructpix2pix,RePaint}, we edit the anchor foreground directly in latent space, avoiding decoding and re-encoding. A numerically aligned anchor alone does not ensure instance persistence, so we also use its instances as explicit references during video generation.

Based on this insight, we propose \ours (Observe, Think, and Carry), a training-free framework (\cref{fig:overview}). Observe identifies instances by matching self-attention clusters to noun regions localized by cross-attention. Think diagnoses numerical errors, applies atomic additions or removals, and verifies each edit by recounting before acceptance, without a language model. Copied instances use perspective-based placement and scaling when the geometry is valid~\cite{Hoiem2005}. Carry Guidance (Carry-CFG) combines text guidance with the contrast between predictions conditioned on clean and re-noised anchors, using instance correspondence to allocate anchor guidance and suppress spurious objects. The framework supports both image-conditioned and text-only video backbones.

We evaluate \ours on CountBench with four backbones: three Wan models ranging from 1.3B to 14B parameters and MiniMax H3~\cite{wan2025wan, wan2025wan22, minimax2026h3}. Across these backbones, \ours improves counting accuracy by 4.6 to 15.2 percentage points and temporal count consistency by 2.6 to 11.9 percentage points over the corresponding base configurations, and outperforms our reproduction of NUMINA under matched settings. Our contributions are summarized as follows:
\begin{itemize}[leftmargin=*,noitemsep,topsep=2pt]
    \item We identify a temporal mismatch between instance identification and layout editability in video diffusion, and address it by separating numerically aligned anchor generation from temporal propagation.
    \item We introduce a training-free procedure for constructing a numerical anchor latent. Observe identifies instances from attention, while Think diagnoses numerical errors, applies local edits, and verifies each edit by recounting.
    \item We propose Carry Guidance (Carry-CFG), an instance-aware classifier-free guidance mechanism that contrasts predictions conditioned on clean and re-noised versions of the anchor. It combines this contrast with text guidance and uses attention-based instance correspondence to weight anchor guidance and suppress spurious instances across frames.
\end{itemize}

\section{Related Work}
\label{sec:related}

\subsection{Video Diffusion and Visual Conditioning}

Diffusion transformers (DiTs)~\cite{peebles2023dit} underpin video generators such as CogVideoX~\cite{yang2025cogvideox}, HunyuanVideo~\cite{kong2024hunyuanvideo}, and Wan~\cite{wan2025wan}. These models combine video VAEs with transformer denoisers, with flow matching~\cite{lipman2023flow} adopted in several recent backbones. Recent generators also support image~\cite{wan2025wan22} and multimodal conditioning~\cite{minimax2026h3}. \ours constructs and verifies a numerical anchor latent from the prompt before using it to condition video generation. In causal VAEs such as Wan's~\cite{wan2025wan}, the initial frame has a separate latent representation, enabling anchor construction before the remaining frames are generated.

\subsection{Counting and Numerical Alignment}

Visual counting estimates the number of objects in observed images through density regression~\cite{ranjan2021famnet}, open-set detection~\cite{liu2024groundingdino}, or text-guided open-vocabulary models such as RT-Counter~\cite{ma2026rtcounter} and MambaCount~\cite{ma2026mambacount}. Numerical alignment in generation further requires controlling the number of synthesized instances. For text-to-image generation, YOLO-Count~\cite{zeng2025yolocount} learns a differentiable counter for gradient-based guidance, while Gu \etal~\cite{gu2026mitigating} adjust the initial noise and optimize early attention to counter noise-induced layout priors. Both methods address static images.

For video, NUMINA~\cite{sun2026numina} extracts countable latent layouts from self- and cross-attention, refines them using spatial and temporal costs, and guides regeneration through cross-attention modulation. Our latent instance extraction builds on NUMINA's clustering. We then edit and verify a fully denoised draft anchor latent before using it to guide video generation, separating anchor construction from temporal propagation.

\subsection{Guidance with Visual and Spatial Conditions}

Classifier-free guidance (CFG)~\cite{ho2022cfg} combines conditional and unconditional predictions to strengthen conditioning. InstructPix2Pix~\cite{brooks2023instructpix2pix} composes image and text guidance with separate scales for image editing, while S-CFG~\cite{shen2024scfg} adapts text guidance across semantic regions. For video, History Guidance~\cite{song2025history} combines predictions conditioned on different history windows or noise levels. Its fractional variant uses partially noised history. Carry-CFG adapts this composition of predictions to numerical control. It contrasts predictions conditioned on clean and re-noised numerical anchors, combines this contrast with text guidance, and allocates stronger anchor guidance to instance regions using attention-based correspondence across frames.

\section{Method}
\label{sec:method}

\begin{figure*}[t]
  \centering
  \includegraphics[width=\textwidth]{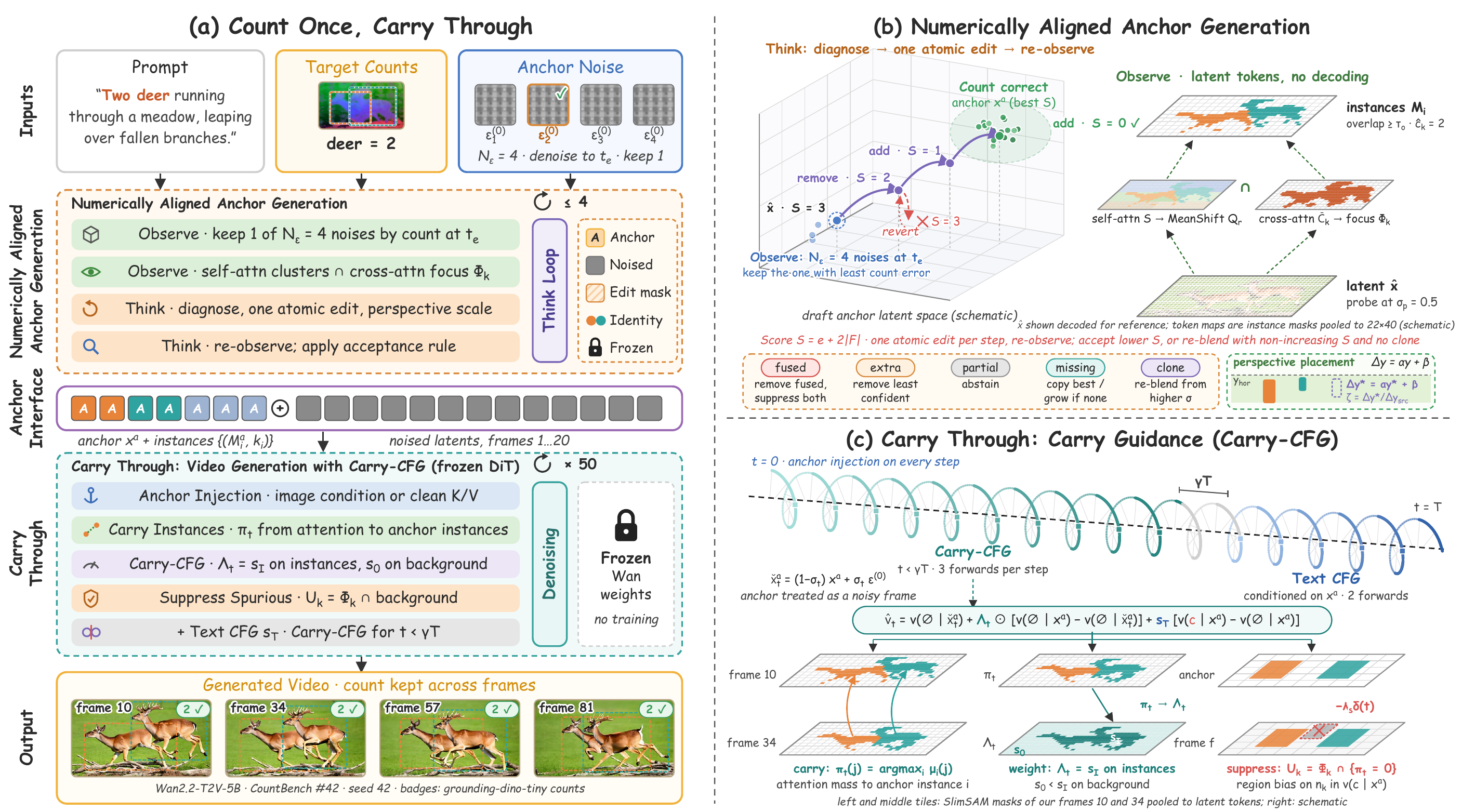}
  \caption{\textbf{Overview of \ours.}
  (a)~For numerically aligned anchor generation, Observe selects a draft latent and extracts its instances, while Think edits and recounts them. Carry then generates the video with Carry-CFG.
  (b)~Instance masks combine self-attention clusters with noun cross-attention. Think follows Select Latent, Spatial Add/Delete, and Identify to choose an edit region, modify its instances, and verify the result.
  (c)~Carry-CFG contrasts predictions conditioned on clean and re-noised versions of the numerical anchor latent. Attention-derived correspondence assigns stronger guidance to instances, while regional suppression discourages extra instances. Guidance is applied during early denoising.}
  \label{fig:overview}
\end{figure*}

\ours separates numerically aligned anchor generation from video generation (\cref{fig:overview}).
Observe and Think construct a numerical anchor latent and its instance masks. Carry-CFG uses these instances as explicit references during video denoising, with all backbone weights frozen.

\subsection{Video Generation Architecture}
\label{sec:backbone}

The Wan backbones~\cite{wan2025wan,wan2025wan22} generate videos in a compressed latent space.
A causal video VAE represents a video as $x\in\mathbb R^{C\times F\times h\times w}$, with $C$ channels, $F$ latent frames, and spatial resolution $h\times w$.
The initial video frame has an independent representation $x^{(0)}$, which provides the slot for constructing our numerical anchor latent.
A text encoder maps the prompt $p$ to conditioning embeddings $c$.

The diffusion transformer (DiT) denoiser~\cite{peebles2023dit} processes patch tokens from all latent frames jointly.
Self-attention relates visual tokens across space and time, while cross-attention associates them with text tokens.
At noise level $\sigma$, the flow-matching model predicts a velocity $v_\theta(x_\sigma,\sigma,c)$ from the noisy video latent $x_\sigma$~\cite{lipman2023flow}.
The sampler uses these predictions to iteratively refine the latent from Gaussian noise, and the VAE decoder converts the result into video frames.
These computations provide the two interfaces used by \ours: attention maps for identifying instances and conditional velocity predictions for Carry-CFG.

\subsection{Framework Overview}
\label{sec:overview}

A prompt $p$ specifies nouns $n_1,\ldots,n_K$ and their target counts $N_1,\ldots,N_K$.
We denote the frozen generator by $G_\theta$.
Sampling performs $T$ updates at $t=0,\ldots,T-1$, with noise levels $\sigma_0=1>\cdots>\sigma_T=0$ and terminal latent $x_T$.
The three stages are
\begin{equation}
  \begin{gathered}
    \hat x=G_\theta\big(p,\epsilon^{(0)}_{\kappa^\star};F{=}1\big),\qquad\mathcal I=\operatorname{Observe}(\hat x),\\
    (x^a,\mathcal I^a)=\operatorname{Think}\big(\hat x,\mathcal I,\{N_k\}\big),\\
    x=\operatorname{Carry}(x^a,\mathcal I^a,p).
  \end{gathered}
  \label{eq:overview}
\end{equation}
Here $F=1$ denotes generation of the single-frame draft anchor latent $\hat x$ from the selected noise $\epsilon^{(0)}_{\kappa^\star}$.
Observe extracts its instances $\mathcal I$. Think returns the \emph{numerical anchor latent} $x^a$, selected after editing and verification, together with its instance set $\mathcal I^a=\{(M_i^a,k_i)\}_{i=1}^{I}$.
Each $M_i^a\subseteq\{1,\ldots,L\}$ is a mask on the anchor's $L$ spatial tokens, and $k_i$ identifies its noun.
Carry generates the video latent $x$ conditioned on $x^a$ and $\mathcal I^a$.

\subsection{Numerically Aligned Anchor Generation}
\label{sec:anchor_construction}

\paragraph{Observe.}
Following the noise-layout connection in~\cite{gu2026mitigating}, we denoise $N_\epsilon$ noise candidates for the draft anchor latent to a screening step $t_e$ and select
\begin{equation}
  \kappa^\star=\arg\min_{\kappa}\ \sum_{k=1}^{K}\big|\hat c_{k,\kappa}-N_k\big|,
  \label{eq:noise}
\end{equation}
where $\hat c_{k,\kappa}$ is the attention-based count estimate for noun $n_k$ in candidate $\kappa$ at $t_e$.
We restart from the selected initial noise and run the full denoising schedule to obtain the draft anchor latent $\hat x$.
We re-noise this completed draft to a probe level $\sigma_p$:
\begin{equation}
  \hat x_{\sigma_p}=(1-\sigma_p)\,\hat x+\sigma_p\,\hat\epsilon,
  \label{eq:probe}
\end{equation}
with fixed Gaussian noise $\hat\epsilon$.
One conditional forward pass yields the noun cross-attention responses $C_k\in\mathbb R^L$ and self-attention affinity $S\in\mathbb R^{L\times L}$, averaged over probe layers and heads, without decoding or an external detector.
For a noun spanning multiple text tokens, $C_k$ also averages their attention probabilities.
Candidate screening reads attention directly from the partially denoised latent. We use these count estimates to rank candidates, whereas editing uses attention probed from the completed draft.

We cluster tokens using self-attention features and spatial positions to obtain candidate instances $\{Q_r\}$~\cite{sun2026numina}.
We subtract the mean row from $S$ and compute a truncated SVD. The three leading components, min-max normalized per coordinate, provide appearance features for each token.
Appending scaled two-dimensional grid coordinates gives the features used by MeanShift clustering, combining affinity similarity with spatial proximity.
To extend noun responses beyond salient object parts, we propagate them through self-attention~\cite{shen2024scfg}:
\begin{equation}
  \bar C_k=\frac{1}{N_p}\sum_{n=1}^{N_p}S^{n}C_k,\qquad
  \tilde C_k=\frac{\bar C_k}{\max_u\bar C_k(u)},
  \label{eq:propagate}
\end{equation}
where $N_p$ is the number of propagation steps.
We threshold $\tilde C_k$ relative to its peak and apply DBSCAN to the selected token positions, removing scattered responses to obtain the noun focus mask $\Phi_k$.
A cluster $Q_r$ is assigned to noun $n_k$ when
\begin{equation}
  o_{rk}=\frac{|Q_r\cap\Phi_k|}{|Q_r|}\ \ge\ \tau_o,
  \label{eq:assign}
\end{equation}
with overlap threshold $\tau_o$.
If several nouns qualify, we choose the one with the highest mean response over $Q_r$, so each cluster is counted at most once.
The assigned masks and labels give $\hat c_k=|\{i:k_i=k\}|$. We use the mean noun response as the instance confidence.

\paragraph{Think.}
Think repeats three steps: \emph{Select Latent}, \emph{Spatial Add/Delete}, and \emph{Identify}.

\noindent\emph{(i) Select Latent.}
Given Observe's masks and counts, we evaluate the current draft anchor latent by its numerical error and a fusion penalty:
\begin{equation}
  \mathcal S=e+w_f\,|\mathcal F|,\qquad e=\sum_{k=1}^{K}\big|\hat c_k-N_k\big|,
  \label{eq:score}
\end{equation}
where $\mathcal F$ contains instances with more than half of their tokens in the focus mask of another under-counted noun, and $w_f>1$ allows their removal before restoring missing instances.
From this diagnosis, we select an instance or spatial region for an atomic edit $a$ with target mask $\Omega_a$.
We prioritize edits for fused, excess, missing, and cloned instances, in that order.
For an excess count, we choose the least confident eligible instance. For a deficit, we use the most confident instance of that noun as a copy source, or grow a new instance when no source exists.
Partial instances are identified by feature similarity to cluster centers and excluded from editing. Extra instances whose boxes lie inside a larger box for the same noun are excluded from removal.

\noindent\emph{(ii) Spatial Add/Delete.}
We remove fused or excess instances and add missing instances by copying a high-confidence instance or growing one from noise.
Copies use perspective-based placement and scaling, or an unscaled fallback for invalid geometry. Source-free growth uses a circular template.
Let $\bar x$ be the current draft anchor latent and $\bar\Omega_a$ the dilation of the selected edit mask $\Omega_a$.
The known latent is $\tilde x=\bar x$, with the source content pasted into the target region for copy edits.
Starting from an edit-dependent step $t_0$, masked regeneration applies
\begin{equation}
  \begin{aligned}
    x_{t_0}&=(1-\sigma_{t_0})\,\tilde x+\sigma_{t_0}\,\hat\epsilon,\\
    x_{t+1}&=\bar\Omega_a\odot\Psi\big(x_t,\hat v^{\,b_a}_t,t\big)+(1-\bar\Omega_a)\odot\tilde x_{t+1},
  \end{aligned}
  \label{eq:regen}
\end{equation}
where $\Psi$ is the sampler update, $\tilde x_{t+1}=(1-\sigma_{t+1})\tilde x+\sigma_{t+1}\hat\epsilon$, and $\hat v^{\,b_a}_t$ is the text-guided velocity with regional bias $b_a$ suppressing removed nouns or amplifying inserted ones.
Masks broadcast over latent patches and channels. Outside the mask, the known content is restored at the current noise level~\cite{RePaint}.

\noindent\emph{(iii) Identify.}
We run Observe on $x'=x_T$ to recover its instance masks and category labels, then recount the instances to obtain the updated score $\mathcal S'$.
Recounting evaluates all nouns, including those outside the edited region, using the same probe settings and noise as before the edit.
The edit is accepted if
\begin{equation}
  \mathcal S'<\mathcal S\ \lor\ \big(a=\text{re-blend}\ \land\ \mathcal S'\le\mathcal S\ \land\ \lnot\operatorname{clone}(x')\big).
  \label{eq:accept}
\end{equation}
Here $\operatorname{clone}(x')$ flags a copy whose latent remains nearly identical to its source. Re-blending regenerates it from a higher noise level.
Otherwise, we restore $\bar x$ and try another candidate edit.
The rejected edit type and noun are excluded for the current state, so the next attempt explores an alternative.
The loop stops when the numerical and clone checks pass, or when its candidate, rejection, or editing budget is exhausted.
The editing budget counts every proposed edit, including rejected proposals.
We select the numerical anchor latent from the visited-state set $\mathcal V$, which includes the initial draft and all evaluated edit results, including rejected proposals:
\begin{equation}
  x^a=\arg\min_{x'\in\mathcal V}\big(\mathcal S(x'),\ \operatorname{clone}(x')\big).
  \label{eq:anchor}
\end{equation}
The comparison is lexicographic by score and then clone flag. Acceptance governs which state is used for subsequent editing.
If editing ends with residual numerical error, Carry receives the best available state under this criterion.

\subsection{Instance-Aware Carry-CFG}
\label{sec:carry}

A verified numerical anchor latent does not by itself ensure that its instances persist during joint video denoising.
Carry-CFG strengthens their influence by contrasting predictions conditioned on the clean anchor latent and a re-noised version, with guidance allocated through instance correspondence.

\paragraph{Clean and re-noised anchor conditions.}
Let $v_t(c'\mid a)$ denote the backbone's velocity prediction on frames $f\ge1$ at sampling step $t$, with text condition $c'$ and anchor condition $a$, where $c$ and $c_\varnothing$ denote the prompt and empty-prompt embeddings.
Models with native image conditioning use the numerical anchor latent $x^a$ through their conditioning interface.
The anchor slot remains fixed at $x_t^{(0)}=x^a$ in both interfaces. Text-only models restore it after each sampler update.
When a shared timestep is used, we additionally cache the anchor's self-attention keys and values at zero noise, once per text-conditioning branch before video denoising.
Each clean-anchor prediction substitutes these cached features for the anchor slot before positional encoding, while the remaining frames retain their current keys and values.
This supplies a clean reference despite the shared timestep.

The contrast branch replaces the clean reference with
\begin{equation}
  \breve x^a_t=(1-\sigma_t)\,x^a+\sigma_t\,\epsilon^{(0)},
  \label{eq:prefix}
\end{equation}
where $\epsilon^{(0)}$ is fixed Gaussian noise for the anchor slot.
For text-only models, this branch processes the re-noised anchor latent without the clean-anchor cache.
All branches share the remaining-frame latent. The re-noised anchor is a temporary contrast condition that leaves the stored anchor fixed.

\paragraph{Joint anchor and text guidance.}
Carry-CFG combines the anchor contrast with text guidance~\cite{ho2022cfg,brooks2023instructpix2pix}:
\begin{equation}
  \begin{aligned}
    \hat v_t={}&v_t(c_\varnothing\mid\breve x^a_t)\\
    &+\Lambda_t\odot\big[v_t(c_\varnothing\mid x^a)-v_t(c_\varnothing\mid\breve x^a_t)\big]\\
    &+s_T\big[v_t(c\mid x^a)-v_t(c_\varnothing\mid x^a)\big],
  \end{aligned}
  \label{eq:carrycfg}
\end{equation}
where $s_T$ is the text guidance scale and $\Lambda_t$ assigns anchor guidance per spatial token.
The second term strengthens the effect of the clean anchor.
The anchor contrast varies the reference under empty text. The text contrast varies the prompt with the clean anchor fixed.
Equivalently, Carry-CFG adds $(\Lambda_t-1)\odot[v_t(c_\varnothing\mid x^a)-v_t(c_\varnothing\mid\breve x^a_t)]$ to ordinary anchor-conditioned CFG.
Without regional suppression, $\Lambda_t=1$ recovers ordinary CFG with the clean anchor, while larger values amplify the anchor contrast.

\paragraph{Instance correspondence and guidance weights.}
We allocate this additional guidance using attention from later frames to anchor instances.
Let $\bar A^{(t)}_{(f,j),(0,u)}$ be the head-averaged self-attention at tracking layer $\ell_p$, from token $j$ in latent frame $f\ge1$ to anchor token $u$.
The anchor background is $M_0^a=\{1,\ldots,L\}\setminus\bigcup_{i=1}^{I}M_i^a$.
Attention mass defines a correspondence to an instance or the background:
\begin{equation}
  \begin{aligned}
    \mu_i^{(t)}(f,j)&=\sum_{u\in M_i^a}\bar A^{(t)}_{(f,j),(0,u)},\\
    \pi_t(f,j)&=\arg\max_{i\in\{0,\ldots,I\}}\mu_i^{(t)}(f,j).
  \end{aligned}
  \label{eq:track}
\end{equation}
We set
\begin{equation}
  \Lambda_t(f,j)=
  \begin{cases}
    s_I, & \pi_t(f,j)\ge1,\\
    s_0, & \pi_t(f,j)=0,
  \end{cases}
  \label{eq:lambda}
\end{equation}
with $s_I>s_0$, broadcasting each weight over the token's channels and spatial patch.
Instance tokens therefore receive stronger anchor guidance than the background.
The anchor masks define reference token sets. Each later-frame token is assigned using its own attention to these sets.
The resulting regions can occupy different spatial positions from the anchor instances, and assignments can vary with both frame and denoising step.
Guidance weights thus follow the current attention correspondence as the video state evolves.
With $s_0=1$, background tokens receive no additional anchor-contrast term, leaving greater flexibility for scene and camera motion.
Using a uniform $\Lambda_t$ without regional suppression gives frame-level anchor guidance.

\paragraph{Suppressing extra instances.}
Let $\Phi_k^{(t)}$ be the spatially filtered noun-attention focus in the later frames, extracted using the Observe procedure.
Tokens that respond to noun $n_k$ but correspond to anchor background form a candidate spurious region:
\begin{equation}
  \mathcal U_k^{(t)}=\big\{(f,j)\in\Phi_k^{(t)}:\pi_t(f,j)=0\big\}.
  \label{eq:spurious}
\end{equation}
For these tokens, we reduce the cross-attention logits to noun $n_k$ in the conditional prediction $v_t(c\mid x^a)$ using a negative regional bias~\cite{sun2026numina}.
Writing $q=(f,j)$, let $\mathcal T_k$ contain the prompt-token indices of noun $n_k$. Its suppression contribution is
\begin{equation}
  b^{(k)}_{qm}(t)=-\lambda_s\,\delta(t)\,
  \mathbf{1}\{q\in\mathcal U_k^{(t)},\ m\in\mathcal T_k\},
  \label{eq:carry_bias}
\end{equation}
where $\lambda_s>0$ controls suppression strength and $\delta(t)$ is a decaying schedule.
This bias targets noun-responsive regions assigned to anchor background, complementing the spatial weighting of the anchor contrast.
At each step, we first collect attention to compute the correspondence and noun-focus regions, then form the suppression regions and their bias. The bias is applied only to layer or prediction-branch computations that have not yet been evaluated.

\paragraph{Guidance schedule.}
We apply Carry-CFG and regional suppression only for $t<\gamma T$, when guidance can influence the emerging layout.
Later steps use ordinary text guidance conditioned on the anchor:
\begin{equation}
  \hat v_t=v_t(c_\varnothing\mid x^a)+s_T\big[v_t(c\mid x^a)-v_t(c_\varnothing\mid x^a)\big],
  \label{eq:late}
\end{equation}
retaining the reference without the extra contrast that could constrain fine motion.
Active Carry-CFG requires three velocity predictions; subsequent anchor-conditioned text guidance uses two.

\begingroup
\raggedbottom
\setlength{\parskip}{0pt}
\setlength{\intextsep}{4pt plus 1pt minus 1pt}
\makeatletter
\renewcommand{\paragraph}{\@startsection{paragraph}{4}{\z@}{4pt}{-1em}{\normalfont\normalsize\bfseries}}
\makeatother
\section{Experiments}
\label{sec:experiments}

\begin{figure}[t]
  \centering
  \captionsetup{skip=4pt}
  \captionsetup[subfigure]{skip=2pt}
  \begin{subfigure}[b]{0.92\columnwidth}
    \centering
    \includegraphics[width=\linewidth]{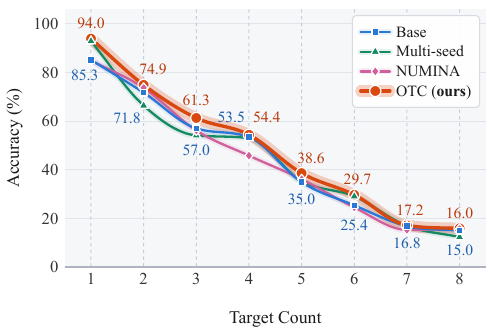}
    \caption{Target number of objects.}
    \label{fig:per_count_acc}
  \end{subfigure}\par\vspace{3pt}
  \begin{subfigure}[b]{0.92\columnwidth}
    \centering
    \includegraphics[width=\linewidth]{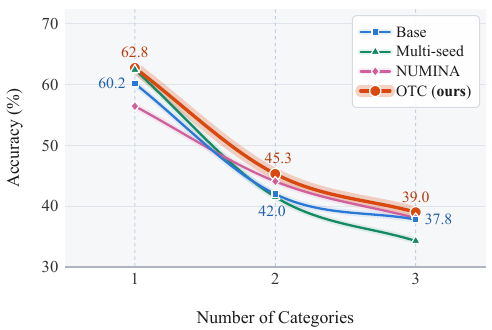}
    \caption{Number of categories in the prompt.}
    \label{fig:per_category_acc}
  \end{subfigure}
  \caption{\textbf{Accuracy by target number and category count on CountBench.} Results are pooled over the evaluated backbones, with groups defined by the numerical specifications in the prompts. The panels use different vertical scales.}
  \label{fig:numerical_complexity}
\end{figure}

\subsection{Experimental Setup}
\label{sec:exp_setup}

\paragraph{Datasets and metrics.}
We evaluate on CountBench~\cite{sun2026numina} (210 prompts) and the generative numeracy subset of T2V-CompBench~\cite{sun2025t2vcompbench} (100 prompts). We use a disjoint set of 40 prompts for hyperparameter validation. All benchmark and development-set metrics use all 81 decoded frames. Grounding DINO~\cite{liu2024groundingdino} provides per-category counts. \textbf{CountAcc} measures exact target-count agreement, averaged over categories and frames. \textbf{TC} measures adjacent-frame count agreement, irrespective of correctness. \textbf{Correct \& Stable (C\&S)} measures the fraction of (video, category) pairs correct throughout all 81 frames. \textbf{All-Category Accuracy (All-Cat.)} measures the fraction of frames satisfying all target counts. Counting metrics are percentages. \textbf{CLIP} reports the mean frame--prompt cosine similarity computed with ViT-B/32~\cite{radford2021clip}, scaled by 100. Gains are computed before display rounding.

\paragraph{Implementation.}
We evaluate four frozen backbones: Wan2.1-T2V-1.3B, Wan2.2-T2V-5B, Wan2.1-T2V-14B~\cite{wan2025wan,wan2025wan22}, and MiniMax H3~\cite{minimax2026h3}. Wan uses 50 UniPC~\cite{UniPC} steps, with Carry-CFG active during the first 30\%. H3 uses the pruned INT8 ConvRot FL2VA checkpoint with a LightX2V Turbo LoRA and 12 steps. \ours uses neither an auxiliary language model nor a pixel-space detector during generation; Grounding DINO is used for evaluation and seed-search baselines.

\paragraph{Baselines.}
We compare with the default sampler (\textbf{Base}) and \textbf{NUMINA}~\cite{sun2026numina}, reproduced using its official implementation and default settings. \Cref{tab:main_results} reports our matched Wan runs.

\subsection{Quantitative Results}
\label{sec:exp_comparison}

\begin{table}[!htbp]
\centering
\footnotesize
\captionsetup{skip=4pt}
\caption{\textbf{CountBench comparison on Wan.} All results are from our own runs with shared settings, seeds, and hardware within each backbone. Parentheses show changes from the corresponding Base (CountAcc/TC: percentage points). CLIP is scaled by 100. Best scores are \textbf{bold}.}
\label{tab:main_results}
\paperTableStyle
\setlength{\tabcolsep}{3pt}
\renewcommand{\arraystretch}{1.0}
\begin{tabular}{lccc}
\toprule
\rowcolor{tableHeader}
\textbf{Method} & \textbf{CountAcc} $\uparrow$ & \textbf{TC} $\uparrow$ & \textbf{CLIP} $\uparrow$ \\
\midrule
\rowcolor{tableGroup}
\multicolumn{4}{l}{\textit{\textbf{Wan2.1-T2V-1.3B}} (81 frames, $832\times480$)} \\
Base & 41.1 & 85.3 & 34.3 \\
\quad + NUMINA & 42.2\,{\scriptsize$(+1.1)$} & 83.7\,{\scriptsize$(-1.6)$} & 34.4\,{\scriptsize$(+0.1)$} \\
\rowcolor{tableOurs}
\quad \textbf{+ \ours} & \textbf{45.7}\,{\scriptsize$(+4.6)$} & \textbf{90.8}\,{\scriptsize$(+5.5)$} & \textbf{35.5}\,{\scriptsize$(+1.2)$} \\
\midrule
\rowcolor{tableGroup}
\multicolumn{4}{l}{\textit{\textbf{Wan2.2-T2V-5B}} (81 frames, $1280\times704$)} \\
Base & 45.2 & 86.4 & 33.8 \\
\quad + NUMINA & 45.9\,{\scriptsize$(+0.7)$} & 86.4\,{\scriptsize$(+0.0)$} & 34.1\,{\scriptsize$(+0.3)$} \\
\rowcolor{tableOurs}
\quad \textbf{+ \ours} & \textbf{51.9}\,{\scriptsize$(+6.7)$} & \textbf{89.0}\,{\scriptsize$(+2.6)$} & \textbf{34.9}\,{\scriptsize$(+1.1)$} \\
\midrule
\rowcolor{tableGroup}
\multicolumn{4}{l}{\textit{\textbf{Wan2.1-T2V-14B}} (81 frames, $1280\times720$)} \\
Base & 54.3 & 84.3 & 34.3 \\
\quad + NUMINA & 58.6\,{\scriptsize$(+4.3)$} & 88.3\,{\scriptsize$(+4.0)$} & 34.3\,{\scriptsize$(+0.0)$} \\
\rowcolor{tableOurs}
\quad \textbf{+ \ours} & \textbf{63.7}\,{\scriptsize$(+9.4)$} & \textbf{90.9}\,{\scriptsize$(+6.6)$} & \textbf{35.8}\,{\scriptsize$(+1.5)$} \\
\bottomrule
\end{tabular}
\end{table}

\begin{table}[!htbp]
\centering
\small
\captionsetup{skip=4pt}
\caption{\textbf{MiniMax H3 on CountBench} using the 12-step distilled, pruned INT8 ConvRot FL2VA configuration. Best results are \textbf{bold}.}
\label{tab:h3_results}
\paperTableStyle
\setlength{\tabcolsep}{3.5pt}
\resizebox{\columnwidth}{!}{%
\begin{tabular}{lccccc}
\toprule
\rowcolor{tableHeader}
\textbf{Method} & \textbf{CountAcc} $\uparrow$ & \textbf{TC} $\uparrow$ & \textbf{C\&S} $\uparrow$ & \textbf{All-Cat.} $\uparrow$ & \textbf{CLIP} $\uparrow$ \\
\midrule
Original prompt & 26.7 & 63.8 & 23.0 & 18.0 & 30.2 \\
Prompt enhancement & 38.2 & 75.1 & 34.8 & 28.1 & 31.3 \\
Seed search & 34.2 & 69.0 & 32.8 & 26.1 & 31.5 \\
NUMINA & 39.4 & 72.8 & 36.0 & 29.6 & 32.2 \\
\rowcolor{tableOurs}
\textbf{\ours} & \textbf{41.9} & \textbf{75.6} & \textbf{38.7} & \textbf{33.0} & \textbf{32.6} \\
\bottomrule
\end{tabular}%
}
\end{table}

\begin{samepage}
\paragraph{CountBench.}
Across the three Wan backbones, \ours improves CountAcc over NUMINA by 3.5, 6.0, and 5.1 percentage points, respectively, while also increasing TC and CLIP (\cref{tab:main_results}). The 14B backbone gains 9.4 points over Base, showing that numerical control remains beneficial with a larger generator. On MiniMax H3, \ours exceeds NUMINA by 2.5 points in CountAcc and 2.8 in TC, and leads all five metrics (\cref{tab:h3_results}). Prompt enhancement comes within 0.5 points of \ours in TC but remains 3.7 points lower in CountAcc. Across the four configurations, equal-weight gains over the corresponding bases average 9.0 points in CountAcc and 6.6 in TC.
\end{samepage}

\begin{figure*}[t]
  \centering
  \includegraphics[width=\textwidth]{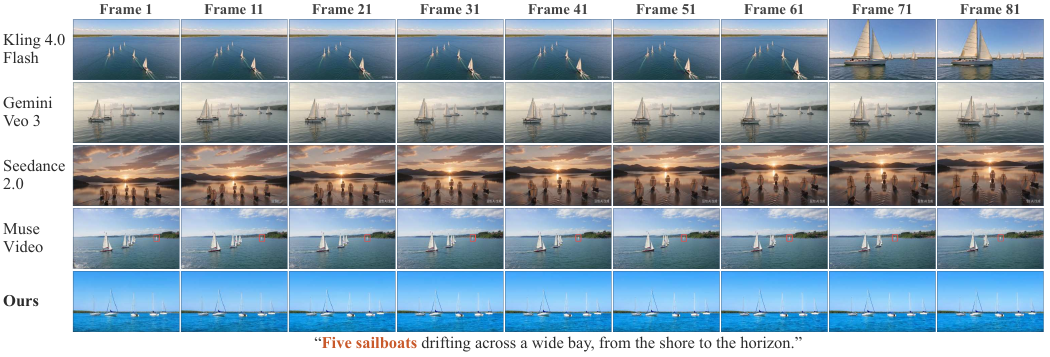}
  \captionsetup{skip=4pt}
  \caption{\textbf{Qualitative comparison on the five-sailboat development prompt (\texttt{dev\_006}).} \ours uses Wan2.1-T2V-1.3B. Nine frames are shown at ten-frame intervals from the first 81 frames, or the last 81 for Meta Muse Video (source frames 160--240). Red boxes mark extra sailboats; all views are cropped to 16:9.}
  \label{fig:sailboats}
\end{figure*}

\paragraph{Effect of numerical complexity.}
\Cref{fig:numerical_complexity} breaks down accuracy by the target number of objects and the number of categories specified in the prompt. Aggregated across the evaluated backbones, \ours achieves the highest accuracy in every plotted group. Its accuracy nevertheless decreases as more objects or categories are requested, highlighting the remaining difficulty of numerical control in these settings.

\begin{table}[!htbp]
\centering
\small
\captionsetup{skip=4pt}
\caption{\textbf{T2V-CompBench generative numeracy} on Wan2.1-T2V-1.3B. Best results are \textbf{bold}.}
\label{tab:t2vcompbench}
\paperTableStyle
\setlength{\tabcolsep}{3.5pt}
\resizebox{\columnwidth}{!}{%
\begin{tabular}{lccccc}
\toprule
\rowcolor{tableHeader}
\textbf{Method} & \textbf{CountAcc} $\uparrow$ & \textbf{TC} $\uparrow$ & \textbf{C\&S} $\uparrow$ & \textbf{All-Cat.} $\uparrow$ & \textbf{CLIP} $\uparrow$ \\
\midrule
Base & 58.2 & 87.8 & 55.8 & 48.3 & 28.8 \\
NUMINA & 59.2 & 89.1 & 57.1 & 49.8 & 28.4 \\
\rowcolor{tableOurs}
\textbf{\ours} & \textbf{73.2} & \textbf{89.8} & \textbf{69.0} & \textbf{69.4} & \textbf{29.7} \\
\bottomrule
\end{tabular}%
}
\end{table}

\begingroup
\widowpenalty=10000
\paragraph{T2V-CompBench.}
On Wan2.1-T2V-1.3B, \ours improves all five metrics over the base model and NUMINA (\cref{tab:t2vcompbench}). CountAcc increases by 15.0 and 14.0 percentage points, respectively, while All-Category Accuracy improves by 21.1 and 19.6 points. TC increases by 2.0 points over Base, compared with 15.0 in CountAcc. The larger gains in exact and joint count accuracy show better agreement with the target numbers beyond increased temporal stability.
\par
\endgroup

\begin{table}[!htbp]
\centering
\footnotesize
\captionsetup{skip=4pt}
\caption{\textbf{Development-set comparison} on the 40 hyperparameter-validation prompts with Wan2.1-T2V-1.3B (81 frames, $832\times480$) and shared sampling settings. Time is seconds/video. Best results among the shown methods are \textbf{bold}.}
\label{tab:dev_main}
\paperTableStyle
\setlength{\tabcolsep}{3pt}
\renewcommand{\arraystretch}{1.0}
\begin{tabular}{lcccc}
\toprule
\rowcolor{tableHeader}
\textbf{Method} & \textbf{CountAcc} $\uparrow$ & \textbf{TC} $\uparrow$ & \textbf{CLIP} $\uparrow$ & \textbf{Time (s)} $\downarrow$ \\
\midrule
Base & 57.7 & 85.6 & 31.8 & \textbf{224.7} \\
NUMINA & 59.3 & 86.9 & 32.2 & 413.2 \\
\rowcolor{tableOurs}
\textbf{\ours} & \textbf{73.2} & \textbf{93.4} & \textbf{32.9} & 233.3 \\
\bottomrule
\end{tabular}
\end{table}

\paragraph{Development set.}
On the 40-prompt development set, \ours improves CountAcc by 15.4 points over Base and 13.9 over NUMINA (\cref{tab:dev_main}). It adds 8.6~s per video over Base and is 179.9~s faster than NUMINA.

\paragraph{Qualitative results.}
\Cref{fig:teaser,fig:sailboats} illustrate numerical alignment across video generators. In \cref{fig:sailboats}, we compare \ours with Kling 4.0 Flash~\cite{kling4release}, Veo 3~\cite{Veo3}, Seedance 2.0~\cite{seedance2026official}, and Meta Muse Video~\cite{meta2026musevideo} on the five-sailboat development prompt. Across the displayed frames, \ours shows five boats, while the other selected outputs show extra boats or a changing number of visible boats.

\subsection{Ablations and Efficiency}
\label{sec:exp_ablation}
\label{sec:exp_analysis}

\begin{table}[!htbp]
\centering
\small
\captionsetup{skip=4pt}
\caption{\textbf{Wan2.1-T2V-1.3B ablation} ($832\times480$, 81 frames). +Frame: fixed, unselected and unedited anchor, without K/V injection or Carry-CFG; +Anchor: latent candidate selection; +All: full \ours; +KV: K/V injection replaces Carry-CFG, retaining Observe and Think.}
\label{tab:ablation}
\paperTableStyle
\resizebox{\columnwidth}{!}{%
\begin{tabular}{lccc}
\toprule
\rowcolor{tableHeader}
\textbf{Configuration} & \textbf{CountAcc} (\%) $\uparrow$ & \textbf{TC} (\%) $\uparrow$ & \textbf{CLIP} $\uparrow$ \\
\midrule
Wan2.1-1.3B (Base) & 41.1 & 85.3 & 34.3 \\
\quad + Frame & 39.9 & 85.1 & 33.2 \\
\quad + Anchor & 45.3 & 90.7 & 33.9 \\
\rowcolor{tableOurs}
\quad \textbf{+ All (\ours)} & \textbf{45.7} & 90.8 & \textbf{35.5} \\
\midrule
\quad + KV & 39.7 & \textbf{96.6} & 34.1 \\
\bottomrule
\end{tabular}%
}
\end{table}

\paragraph{Anchor construction and guidance.}
Fixing an unedited anchor (+Frame) reduces CountAcc from 41.1 to 39.9 (\cref{tab:ablation}), so anchoring alone does not establish the target count. Candidate selection (+Anchor) raises CountAcc and TC to 45.3 and 90.7, accounting for most of the gain over Base. Adding Think and Carry-CFG (+All) contributes a further 0.4 and 0.1 points, respectively, and raises CLIP from 33.9 to 35.5. This comparison measures their combined contribution. Replacing Carry-CFG with K/V injection while retaining Observe and Think (+KV) yields 96.6\% TC but only 39.7\% CountAcc, below Base. Thus, stable counts can still be incorrect. Full \ours also uses clean-anchor K/V conditioning; Carry-CFG adds the anchor contrast.

\section{Conclusion}
\label{sec:conclusion}

We presented \ours, a training-free framework in which Observe and Think establish and verify the count in a numerical anchor latent, and Carry propagates it through the video with Carry-CFG. Across four backbones, \ours improves average CountAcc and TC on CountBench by 9.0 and 6.6 percentage points, respectively, and outperforms NUMINA under matched settings. The ablations also show that temporally consistent counts are not necessarily correct, which supports establishing the count first.

\paragraph{Limitations.}
Accuracy still declines as prompts request more objects or categories. The anchor also represents only the initial frame, leaving objects that enter later and time-varying counts to future work.

\endgroup
\clearpage
\section*{LLM Usage Statement}
OpenAI's ChatGPT and Codex were used for language editing, restructuring and drafting manuscript text from author-provided material, related-work research and citation verification, code development, and debugging experimental scripts. They also assisted with refining author-prepared experimental analyses. GPT (configuration: \texttt{GPT 5.6 sol ultra}) was used to generate our 40-prompt development set. Manuscript content was shared with OpenAI for reviewer-style feedback. All AI-assisted code and manuscript content were reviewed and validated by the authors, who take full responsibility for the methodology, experiments, and conclusions.

{
    \small
    \bibliographystyle{ieeenat_fullname}
    \bibliography{main}
}

\end{document}